# Topic Diffusion Discovery Based on Deep Non-negative Autoencoder

Sheng-Tai Huang
*Department of Information Management*
*National Sun Yat-sen University*
Kaohsiung, Taiwan
shengtai.huang28@gmail.com

Yihuang Kang
*Department of Information Management*
*National Sun Yat-sen University*
Kaohsiung, Taiwan
ykang@mis.nsysu.edu.tw

Shao-Min Hung
*Department of Information Management*
*National Sun Yat-sen University*
Kaohsiung, Taiwan
shaominhung96@gmail.com

Bowen Kuo
*Department of Information Management*
*National Sun Yat-sen University*
Kaohsiung, Taiwan
bowenkuo@outlook.com

I-Ling Cheng
*Graduate Institute of Library and Information Science*
*National Chung Hsing University*
Taichung, Taiwan
chengi428@gmail.com

***Abstract*— Researchers have been overwhelmed by the explosion of research articles published by various research communities. Many research scholarly websites, search engines, and digital libraries have been created to help researchers identify potential research topics and keep up with recent progress on research of interests. However, it is still difficult for researchers to keep track of the research topic diffusion and evolution without spending a large amount of time reviewing numerous relevant and irrelevant articles. In this paper, we consider a novel topic diffusion discovery technique. Specifically, we propose using a Deep Non-negative Autoencoder with information divergence measurement that monitors evolutionary distance of the topic diffusion to understand how research topics change with time. The experimental results show that the proposed approach is able to identify the evolution of research topics as well as to discover topic diffusions in online fashions.**

## I. INTRODUCTION

Due to the prosperity of information and communication technology, digital publication has enabled researchers to publish their papers much easier. However, to identify novel research areas has also become unprecedentedly laborious, not to mention how to keep track of the up-to-date research progresses. Although most research papers are well-categorized, the categories are so broad that they still cover abundant papers. Hence, to define research areas, researchers have been using topic modeling algorithms to explore unknown research areas and trends.

Topic modeling algorithms can be mainly distinguished into probabilistic methods and matrix factorization techniques. Probabilistic methods, such as Latent Dirichlet Allocation (LDA) [1], attempts to discover the word distribution for each topic. While matrix factorization algorithms, such as Singular Value Decomposition (SVD) [2] and Non-negative Matrix Factorization (NMF) [3], decompose sparse document-term matrix into a series of dense, low-rank matrices that gather terms/words with similar occurrences into topics. Typical topic modeling algorithms do not consider hierarchical/multi-layer structure, which makes them unable to learn more complex topic-term relationships since topics might be composed of topics, but there are modified matrix factorization algorithms, such as multi-layer/hierarchical NMF [4], which incorporate with the multi-layer structure.

Another problem of typical topic models is that they usually do not take into account topic evolutions and diffusions. All documents and terms are processed simultaneously to generate a document-term matrix for a given set of text corpus, but term frequency (tf) or term frequency-inverse document frequency (tf-idf) actually changes in different time periods. As topics might split or merge over time, rather than building topic models that take bunches of text corpus periodically, we should consider updating topic models incrementally with streaming of text data so that researchers can inspect the change of topics.

In this paper, we propose to adopt the concept of NMF and exploit autoencoders with non-negative constraints for the purpose of building interpretable and expressive topic models that discover the inherent multi-layer topic-term structure of given sets of text corpus in different time slices. Furthermore, we consider training topic models in an online learning manner with a sliding window that processes a series of text corpus as well as utilizing information divergence to evaluate the magnitude of topic diffusions over time.

The rest of this paper is organized as follows. In Section 2, we review topic modeling algorithms and related research. Our proposed approach, Deep Non-negative Autoencoder (DNAE) is discussed in Section 3. In Section 4, we present our experimental results on real-world research articles about machine learning and conclude with our findings in the last section.

## II. BACKGROUND AND RELATED WORK

To discover potential research areas, researchers need to do intensive reviews on large numbers of research articles published by journals, conferences, and other research communities regularly, which makes it difficult for a researcher to keep up with the latest research progress. In our previous work [5], we proposed a topic learning and diffusion discovery algorithm to identify research areas, trace the progress of research, and further help expose unknown research areas. Nonetheless, researchers have proposed algorithms with multi-layer structure for the purpose of capturing complex representations from data in recent years. Moreover, the autoencoders [6], [7] have also become options of building topic models due to their flexible model structure that is able to train models with large text data in online fashion. Here, inspired by the flexibility of the autoencoder, our proposed Deep Non-negative Autoencoder (DNAE) is inherently explainable and able to help understand topic/term evolution in probabilistic manner.

A typical topic model is built with a text corpus consisting of *n* documents and an associated dictionary with m terms, and they are processed to generate an $n \times m$ document-term frequency matrix *X*. This matrix is usually sparse because most terms do not occur in all the documents, which makes researchers unable to identify the relationships among documents and terms. Therefore, researchers usually assume that documents and terms are composed of topics [8], [9], and topic modeling algorithms are used to discover the more intuitive topics.

Topic modeling algorithms can be distinguished into two major genres, probabilistic (e.g. LDA) and matrix factorization (e.g. SVD and NMF). Typical LDA [1] attempts to approximate the latent word distribution in each topic and generates recognizable topics. However, LDA may generate inconsistent results with similar parameter settings, which is impractical in the online applications of topic diffusion discovery. Matrix factorization algorithms, on the other hand, perform low-rank approximation to the input document-term frequency matrix to discover bag-of-words with similar occurrences that represent the latent topics within documents. For instance, the SVD is a popular matrix factorization algorithm used to discover topic-term relationships. However, the orthogonality of the SVD enforces all topics to be completely independent of each other, which violates the fact that topics might be overlapped because topics may share the same terms [10], [11]. Also, the decomposed matrices from the NMF are non-negative and additive, and the topics from these matrices can be expressed as polynomial functions, which helps understand the importance of each portion via its part-based topic-term representations. Typical NMF, as shown in Fig. 1, decomposes the aforementioned document-term matrix *X* into two lower-rank matrices, *W* and *H*, such that

$$X \approx WH \tag{1}$$

where $W$ is $n \times k$ matrix, $H$ is $k \times m$ matrix, $k$ is a hyper-parameter that denotes the number of topics and $k < min(n, m)$. To find a low-rank approximation of document-term matrix *X*, the NMF reaches the goal by minimizing Frobenius norm $\|.\|_F$, as:

$$\min f(W,H) = \frac{1}{2}\|X - WH\|_F^2, s.t. W \geq 0, H \geq 0 \tag{2}$$

where all elements in *W* and *H* are non-negative.

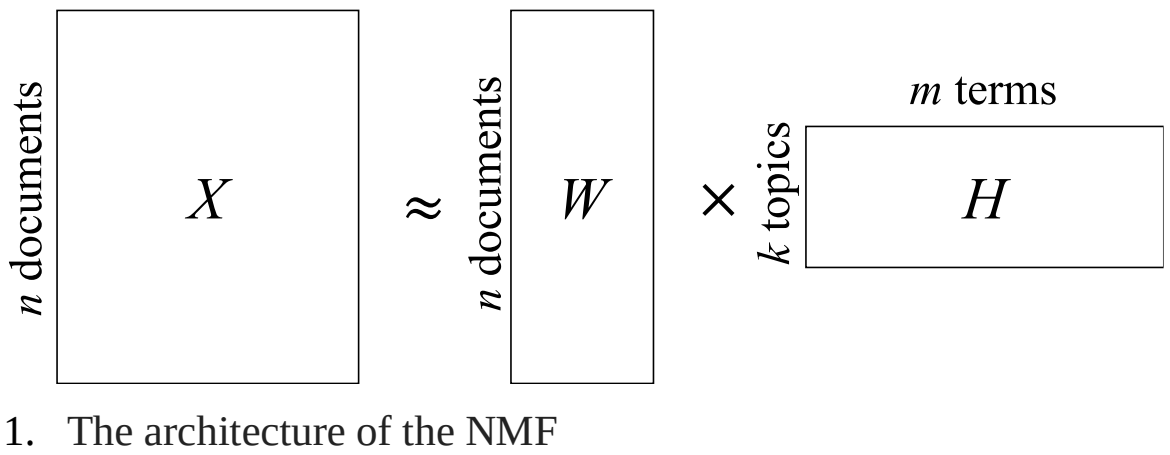

Fig. 1. The architecture of the NMF

A multi-layer structure of NMF has more expressive power to learn complex data representations [4], [12]. The difference between NMF and hierarchical/multi-layer NMF (hNMF) is that hNMF keeps decomposing the matrix (the topic-term matrix in our case) *H* into a lower-rank matrix. For example, NMF decomposes the input matrix *X*, such that $X \approx W_1 H_1$, whereas hNMF, as shown in Fig. 2, then further decomposes $H_1$ so that $X \approx W_1 W_2 H_2$. An *l*-layer hNMF can be expressed as:

$$X \approx W_1 W_2 \dots W_l H_l \tag{3}$$

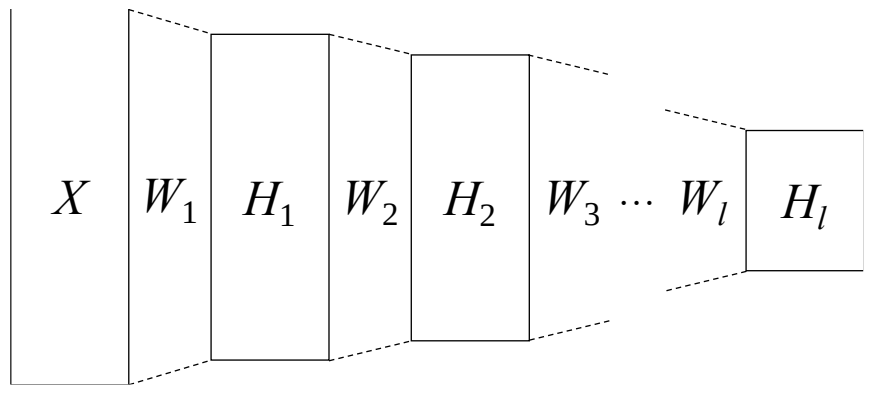

Fig. 2. The architecture of the hNMF

The multi-layer structure of the hNMF-like algorithms may potentially discover term/topic hierarchy as well as other latent relationships/dependency among topics. They are able to learn complicated representations of topics and terms but incapable of dealing with large-scale text datasets when the main memory limit is a concern. Consequently, we borrow the idea from Deep Autoencoder-like Nonnegative Matrix Factorization [13], which exploits gradient-based methods that train the autoencoder with mini-batch sampling and impose non-negative constraints on weights to imitate the hNMF.

As topics keep evolving over time, the components (weighted terms) of topics are constantly changing. In evolutionary NMF (eNMF) [14], the authors argue that the representations of topics from the following time slices are based on the previous ones, and thus it will only have some adjustment on the values in raw matrix and decomposed matrices in different times. However, the size of input matrices from different time slices in eNMF must be fixed. Therefore, we adopt the main concept of the eNMF by training the model in online fashion to retain the weights of the autoencoder in different times. The initial weights of the model in the following time slice are the weights from the pre-trained DNAE model in the previous time slice. In the next section, we indicate the details of our proposed algorithm, DNAE.

## III. Topic Diffusion Discovery based on Deep Non-negative Autoencoder

As discussed previously, our proposed Deep Non-negative Autoencoder (DNAE) is an autoencoder with non-negative constraints imposed on weights to simulate hNMF. Different from the hNMF which generates a series of topic-term matrices in layer-wise training fashion, the DNAE constructs its encoder and decoder by compressing and decompressing the matrix. Fig. 3 illustrates the architecture of the DNAE. Here, $H_1, H_2, \dots, H_l$ and $H_1', H_2', \dots, H_l'$ are the weights of the autoencoder, whereas $W_1, W_2, \dots, W_l$ and $W_1', W_2', \dots, W_l'$ represent the intermediate output matrices. $H_i$ is a non-negative coefficient matrix used to encode/decode the data matrix *X*. The data matrix *X* is encoded into intermediate, lower-dimensional, latent data representation matrix $W_i$.

To simulate the hNMF, we propose to remove the bias and nonlinear activation function in each layer so as to retain the intermediate weights recoverable to the original input matrix. Here, the goal of the DNAE is to minimize the reconstruction error between the input document-term matrix *X* and the reconstructed matrix *X'*.

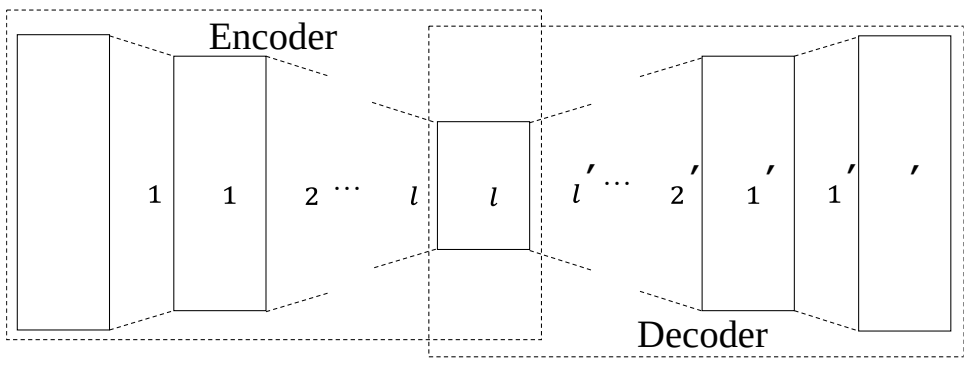

Fig. 3. The architecture of the DNAE

As shown in Fig. 4, we train a DNAE and continuously update its weights with new text data. In NMF-like topic models, Hungarian Algorithm [15] is often used to match topics between time slices. Here, as the DNAEs are trained in online fashions that weights in the network are being updated with text data (articles) in corresponding time slices, the topics are automatically aligned to topics from previous time slices. Although online learning enables DNAEs to adopt learned representations from previous time slices without redundant articles, it may cause catastrophic interference problems [16], which makes the topics in the following time slices more inconsistent with the topics in the previous time slices. Therefore, we combine the document-term matrix of the following time slice with the matrix from the previous one time slice (i.e. the size of sliding window is 2). After training DNAEs, we obtain weights $H_1, H_2, \dots, H_l$ from hidden layers in the encoder part and multiply them into a topic-term matrix *U*. Subsequently, to observe the term diffusions in topics, we normalize the topic-term matrix *U* into relative frequency form in term wise to make the summation of each term equals to 1, which can be regarded as conditional probability of word in topics (i.e. $P(topic_k|term_i)$). In different time slices, we obtain a series of topic-term matrices $U^1, U^2, \dots, U^t$, and the generalized Jensen-Shannon divergence ($D_{GJS}$) [17], [18] is utilized to compute the term diffusions in topics. The $D_{GJS}$ is defined as:

$$D_{GJS}(P_1, P_2, \dots, P_t) = H\left(\sum_{i=1}^{t} \pi_i P_i\right) - \sum_{i=1}^{t} \pi_i H(P_i) \quad (4)$$

where $\pi_i$ is the weight for each discrete probability distribution, and *H*(*x*) is *k*-ary Shannon entropy defined as:

$$H(x) = -\sum_{i=1}^{k} P(x_i) \log_k P(x_i) \quad (5)$$

The $D_{GJS}$ is used to observe topic diffusions in different time slices (e.g. term *i* from $U^t$ and term *i* from $U^{t+1}$). To evaluate the degree of topic diffusion, a statistical significance threshold of the $D_{GJS}$ is used, and the threshold is defined as:

$$D_{GJS|k,t} \cong \frac{\chi^2_{df,1-\alpha}}{2N \ln(k)} \quad (6)$$

where $df = (k-1)(t-1)$ is the degree of freedom, $\alpha$ is the statistical significance level, and *N* is the total number of cells (*k* by *t*) used in calculating the Chi-square statistic $\chi^2$ in different time slices. Note that *k* is the number of topics and *t* is the number of observed time slices/frames. Also, higher $D_{GJS}$ denotes the higher degree of topic diffusion/divergence.

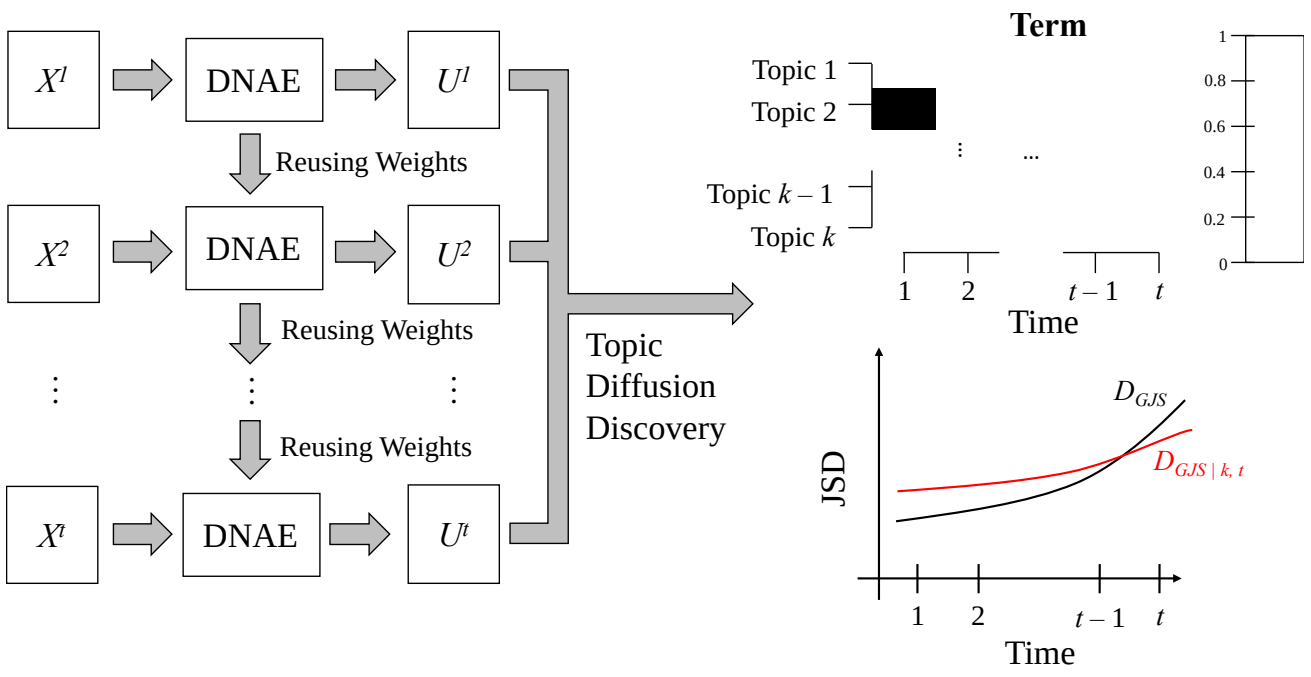

Fig. 4. The entire topic diffusion discovery process

## IV. Experiment and Discussion

In this section, we present our experimental results of the proposed approach and discuss our findings. All of our work was implemented by R 3.5.1 [19] with *Keras* [20], *ggplot2* [21]. To evaluate the feasibility of our approach, we collected 31,904 open-access full-text articles related to Machine Learning in 2007/01-2019/12 from arXiv.org stat.ML [22] (TABLE I). Instead of using common words, an ML-related dictionary with 17,240 terms/words was created here by extracting the author-defined keyword list within all articles. We downloaded the data through arXiv.org API [23], cleaned the data with R *tm* package [24]. Then we removed redundant words and mapped the abbreviation or plural nouns to their corresponding terms. As the ML-related dictionary is used, stemming and stopwords removals are not required. As there are relatively fewer articles from 2007 to 2014, we here consider the period as a time slice in the experiments to build an initial topic model.

TABLE I. THE NUMBER OF ARTICLES 2007 ~ 2019

| Year | # of Articles |
|---|---|
| 2007 ~ 2014 | 4,263 |
| 2015 | 1,665 |
| 2016 | 2,417 |
| 2017 | 3,708 |
| 2018 | 7,968 |
| 2019 | 11,883 |

As discussed previously, each layer in the DNAE has been imposed non-negative constraints, and the biases, nonlinear activation functions are removed to simulate hNMF for model interpretability purposes. RMSE is used to evaluate the reconstruction errors and the model performance. Online learning may cause the aforementioned catastrophic interference problems [16], and thus we consider a 2-year sliding window. The size of the sliding windows is a hyper-parameter and should be selected properly in different online learning applications. As the distribution of terms is skewed and low-tailed, we applied tf-idf transformation for each document-term matrix separately to make the distributions of term frequencies more smooth. In the experiment, we empirically trained DNAE with three hidden layers (number of neurons are 50, 20, 50, respectively), then we extracted the weights from the encoderpart of the DNAE to create 20 by 17,240 topic-term matrix $U$ in each time slice.

Appendix A shows the top-5 keywords of all 20 topics identified by the topic-term matrix $U$ of the DNAE. We can see that most dominant terms in the fields of machine learning do not change significantly. For example, it is clear that topic 4 is mainly about cluster analysis, whereas topic 9 is closely related to the topic modeling techniques.

As there are still changes of term ranking, we provide the observation of terms diffusions in topics for further discussions. The $D_{GJS}$ is used to evaluate the magnitude of the changes. In Fig. 5, we can see topic 1 can be recognized as a topic related to "reinforcement learning", and we know "markov decision process" is mainly used in this field today. Therefore, we may consider the "markov decision process" a "narrow" term as it is not widely applied to other topics in recent decades. Fig. 6 shows another example, "nonnegative matrix factorization", which was related to topic 13 and topic 1 in recent years. The "nonnegative matrix factorization" is not only used as a dimensionality reduction and topic modeling technique but also used in "reinforcement learning" for learning transition matrices lately. Thus, we may say that "nonnegative matrix factorization" is a "broad" term since it has occurred in more than one topic and the $D_{GJS}$ is increasing. On the other hand, "latent dirichlet allocation", as shown in Fig. 7, is usually used in topic modeling, but it is now also applied to "social media" and "news" (weighted terms related to Topic 20) data analysis tasks. In this case, "latent dirichlet allocation" has been discussed in more and more topics, so we may consider it a "divergent" term. Besides, "graph" and "kernel" are both related to "support vector machine" and "graph neural network". As shown in Fig. 8, the terms are changing in topic 11 which may suggest that there are fewer applications on "support vector machine" and more and more discussions on "graph neural network" Fig. 8 also indicates that "graph neural network" has become more and more popular, and we may consider it a "convergent" term.

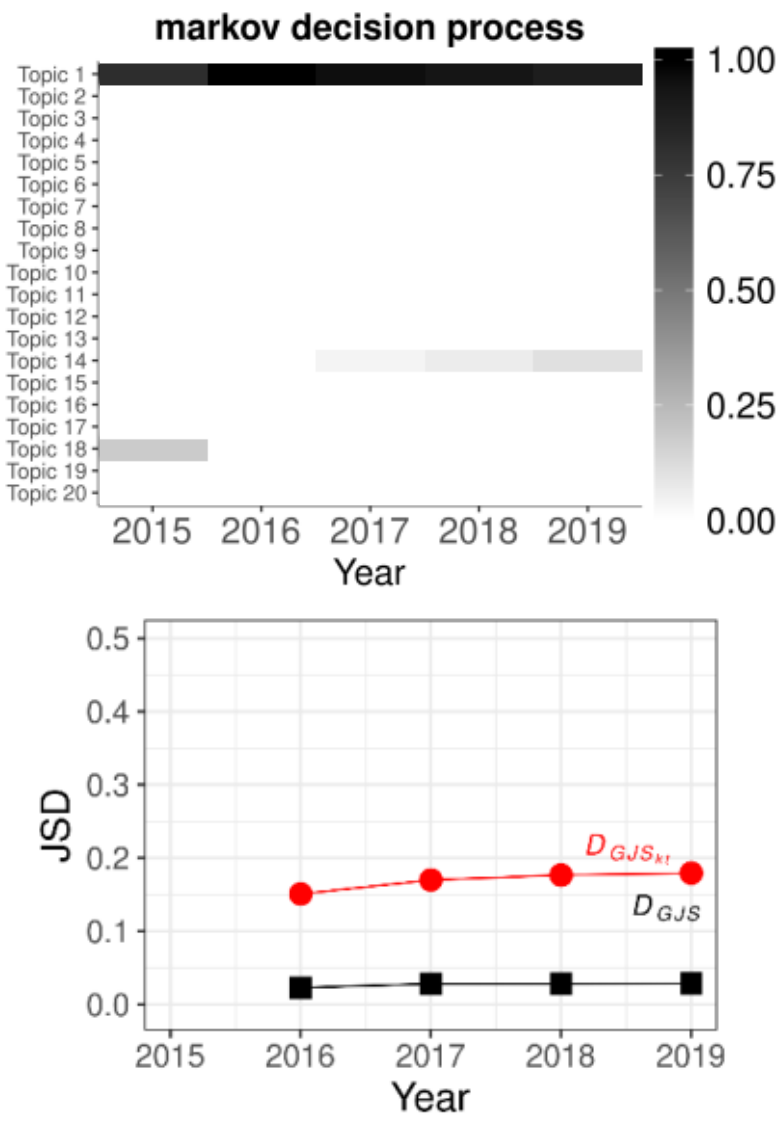

Fig. 5. Narrow term "markov decision process"

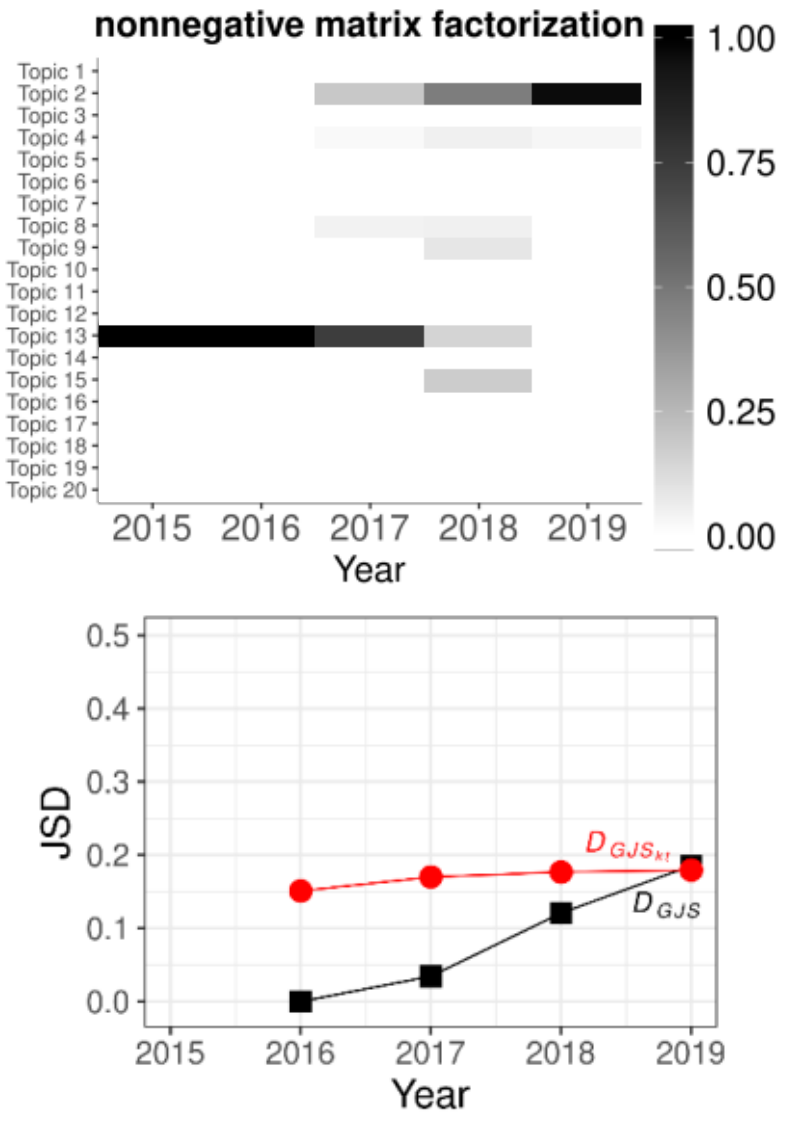

Fig. 6. Broad term "nonnegative matrix factorization"

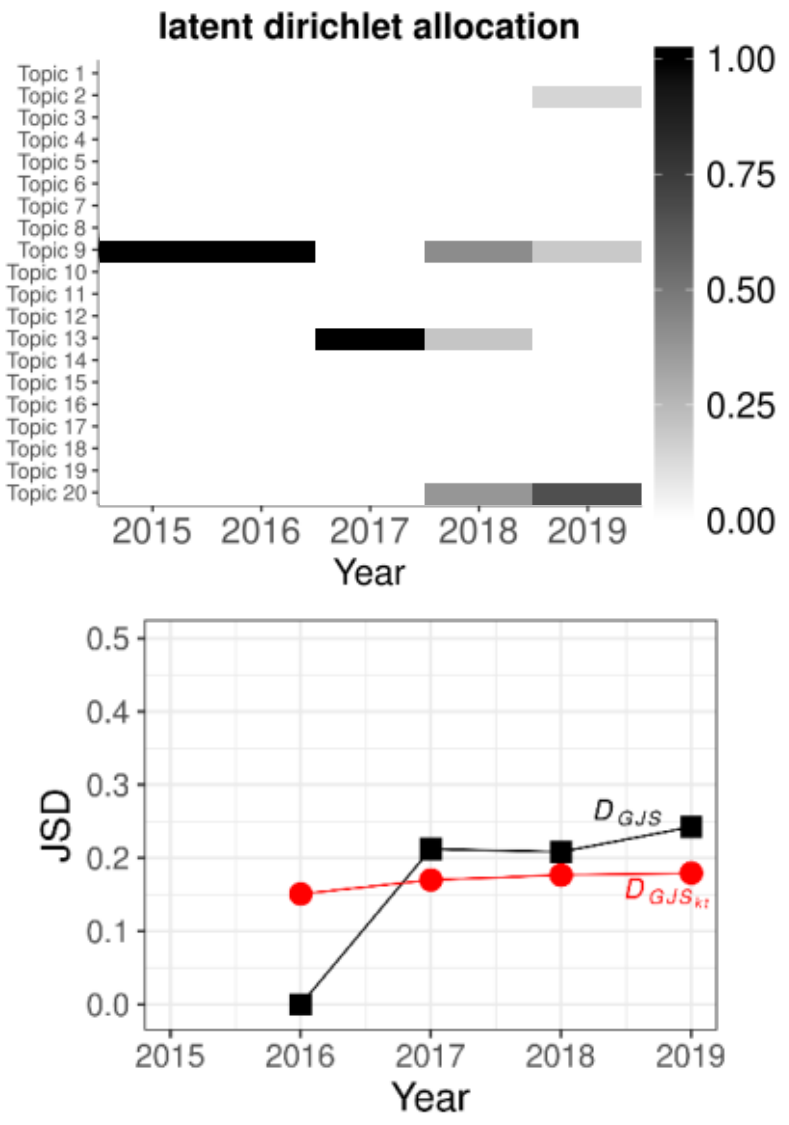

Fig. 7. Divergent term "latent dirichlet allocation"

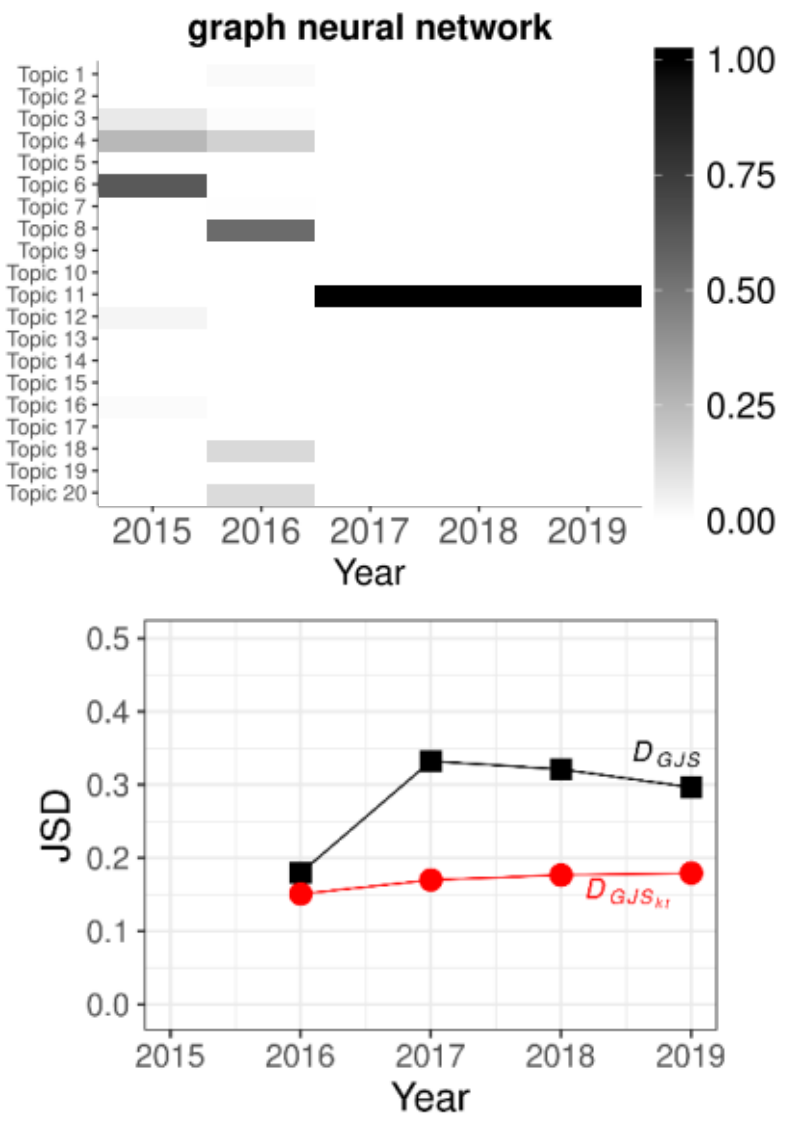

Fig. 8. Convergent term "graph neural network"

## V. CONCLUSION

We proposed a novel topic modeling and diffusion discovery technique that incorporates a multilayer autoencoder with non-negative constraints— a Deep Non-negative Autoencoder (DNAE) that imitates hierarchical NMF but more flexible model architecture able to provide easy-to-understand topics and help keep track of the topic evolutions/diffusions with time. The experiment results show that our approach can discover dynamics of topics as well as to help researchers understand the shifts of research of interests.

## APPENDIX A. TOP-5 KEYWORDS OF ALL 20 TOPICS

| Year / Topic | 2007 ~ 15 | 2015 ~ 16 | 2016 ~ 17 | 2017 ~ 18 | 2018 ~ 19 |
|---|---|---|---|---|---|
| **Topic 1** | policy, markov decision process, value function, reward learning, lwd | policy, markov decision process, reinforcement learning, agent, value function | policy, agent, markov decision process, reinforcement learning, wireheading | policy, agent, reward, reinforcement learning, markov decision process | policy, agent, reward, reinforcement learning, markov decision process |
| **Topic 2** | dictionary, dictionary learning, sparse coding, coding, sparse representation | dictionary, dictionary learning, sparse coding, sparse representation, orthogonal matching pursuit | dictionary, dictionary learning, sparse coding, sparse representation, photometric stereo | dictionary, dictionary learning, sparse coding, sparse representation, orthogonal matching pursuit | dictionary, audio, dictionary learning, nonnegative matrix factorization, source separation |
| **Topic 3** | dropout, recurrent neural network, deep, restricted boltzmann machine, neural network | recurrent neural network, long short term memory, neural network, convolution neural network, machine translation | convolution neural network, long short term memory, recurrent neural network, neural network, machine comprehension | convolution neural network, long short term memory, recurrent neural network, common objects in context, optimal brain damage | pruning, convolution neural network, segmentation, quantization, common objects in context |
| **Topic 4** | cluster, clustering, kmeans clustering, | cluster, clustering, kmeans clustering, | cluster, clustering, kmeans clustering, | cluster, clustering, kmeans clustering, | cluster, clustering, kmeans clustering, |

| Year / Topic | 2007 ~ 15 | 2015 ~ 16 | 2016 ~ 17 | 2017 ~ 18 | 2018 ~ 19 |
|---|---|---|---|---|---|
| | spectral clustering, clustering evaluation | number of clusters, delaunay tessellation | number of clusters, moving object trajectories | clustering algorithm, number of clusters | clustering algorithm, spectral clustering |
| **Topic 5** | lasso, screening, group lasso, high dimensional linear model, group | screening, safe, lasso, safe screening, safe rules | lasso, screening, safe, group lasso, high dimensional linear model | time series, forecasting, lasso, ordinary least squares, treatment | time series, forecasting, long short term memory, recurrent neural network, dynamic time warping |
| **Topic 6** | tree, graph, forest, random forest, statistical causal inference | tree, forest, random forest, classifier, survival | fairness, tree, random forest, classifier, forest | fairness, fair, discrimination, group, fair classification | fairness, fair, group, attribute, counterfactual |
| **Topic 7** | copula, gaussian copula, myoelectric control, fmle, copula density | copula, wasserstein, optimal transportation, transport, wasserstein distance | generative adversarial network, discriminator, adversarial neural networks, generative, adversarial autoencoder | generative adversarial network, discriminator, adversarial neural networks, generative, adversarial setting | generative adversarial network, discriminator, adversarial neural networks, fid, generative |
| **Topic 8** | matrix completion, subspace, low rank, alternating direction method of multiplier, rank minimization | mean absolute percentage error, stochastic convex optimization, rank minimization, convergence rate, stochastic variance reduced gradient | stochastic variance reduced gradient, stochastic gradient descent, stochastic convex optimization, nesterovs smoothing technique, convergence rate | multiplicative perturbations, stochastic variance reduced gradient, stochastic gradient descent, convex, sparse and low rank recovery | convex, sparse and low rank recovery, sparse convex optimization, shift and invert preconditioning, stochastic gradient descent |
| **Topic 9** | topic, latent dirichlet allocation, posterior, dirichlet, variational | topic, posterior, variational, latent dirichlet allocation, topic model | posterior, variational, variational autoencoder, variational inference, latent | posterior, variational, variational autoencoder, latent, variational inference | variational autoencoder, latent, posterior, variational, latent variable |
| **Topic 10** | bandit, bandit problem, multiarmed bandit, reward learning, reward | bandit, bandit problem, reward learning, multiarmed bandit, lwd | bandit, multi armed bandit, wireheading, bandit problem, reward learning | bandit, multi armed bandit, contextual bandit, bandit problem, multi armed bandit algorithms | bandit, multi armed bandit, contextual bandit, thompson sampling, multi armed bandit algorithms |
| **Topic 11** | kernel, support vector machine, classifier, reproducing kernel hilbert space, reproducing kernel | kernel, reproducing kernel hilbert space, canonical correlation analysis, radial basis function, reproducing kernel | graph, kernel, graphlet, laplacian, nonlinear dimensionality reduction | graph, graph convolution neural network, graphlet, laplacian, graph kernel | graph, graph convolution neural network, graph neural network, gegenbauer neural network, graph kernel |
| **Topic 12** | smml, minimum message length, exponential family, step functions, order dependence | adversarial, adversarial example, generative adversarial network, adversarial training, discriminator | adversarial example, attack, adversarial, adversarial perturbation, adversarial training | attack, adversarial example, adversarial, adversarial perturbation, adversarial attack | attack, adversarial example, adversarial, adversarial attack, adversarial perturbation |
| **Topic 13** | nonnegative matrix factorization, matrix factorization, ranking, factorization, hottopixx | nonnegative matrix factorization, matrix factorization, factorization, hottopixx, nonnegative rank | topic, topic model, latent dirichlet allocation, embedding, nonnegative matrix factorization | embedding, word embedding, recommendation, hottopixx, word2vec | embedding, language, word embedding, transformer, bert |
| **Topic 14** | regret, regret bound, exponentially weighted forecaster, communicationfree, game | regret, communicationfree, exponentially weighted forecaster, regret bound, myoelectric control | regret, regret bound, exponentially weighted forecaster, decision theoretic online learning, online convex optimization | regret, regret bound, exponentially weighted forecaster, decision theoretic online learning, predicting with expert advice | regret, regret bound, exponentially weighted forecaster, decision theoretic online learning, online convex optimization |
| **Topic 15** | tensor, tensor decomposition, coupled matrix tensor factorization, | tensor, tensor decomposition, tensor completion, | tensor, tensor completion, tensor decomposition, rank, | tensor, tensor decomposition, tensor completion, rank, | tensor, tensor decomposition, tensor completion, hypergraph, |

| Year / Topic | 2007 ~ 15 | 2015 ~ 16 | 2016 ~ 17 | 2017 ~ 18 | 2018 ~ 19 |
|---|---|---|---|---|---|
| | multivariate tools, bioactivity | coupled matrix tensor factorization, multivariate tools | low rank | low rank | rank |
| **Topic 16** | smml, minimum message length, myoelectric control, order dependence, link indicator kernels | adversarial, adversarial example, generative adversarial network, adversarial training, discriminator | anomaly, anomaly detection, outlier, detection, time series | anomaly, anomaly detection, detection, outlier, outlier detection | anomaly, anomaly detection, detection, outlier, outlier detection |
| **Topic 17** | privacy, differential privacy, privacy preservation, data privacy, local sensitivity | privacy, differential privacy, privacy preservation, false discovery rate, randomized response | privacy, differential privacy, privacy preservation, local sensitivity, sensitivity | privacy, differential privacy, privacy preservation, federated, utility | privacy, differential privacy, federated, federated learning, privacy preservation |
| **Topic 18** | agent, reward learning, bayesian exploration, lwd, corl | game, approachability, reward learning, bayesian exploration, lwd | survival, game, chess, reward learning, churn | tree, forest, random forest, classifier, decision tree | tree, forest, random forest, decision tree, survival |
| **Topic 19** | privacy, myoelectric control, differential privacy, decision under uncertainty, reward learning | myoelectric control, decision under uncertainty, machine learning evaluation, matching selection, edge ai | myoelectric control, decision under uncertainty, machine learning evaluation, matching selection, link indicator kernels | persistence, persistence diagram, persistence landscape, persistent homology, topological data | persistence, persistence diagram, persistent homology, persistence landscape, chatter |
| **Topic 20** | padic, dendrogram, ultrametric, spatial partition, link indicator kernels | padic, anomaly, graphlet, anomaly detection, detection | padic, fake news, twitter, news, spatial partition | twitter, padic, fake news, news, social media | twitter, padic, emotion, sentiment, social media |